\documentclass[11pt,a4paper]{article}
\usepackage[table,dvipsnames]{xcolor} 
\usepackage[hyperref]{emnlp2018}
\usepackage{times}
\usepackage{latexsym}
\usepackage{adjustbox}

\usepackage{url}

\aclfinalcopy 
\def\aclpaperid{2133} 

\usepackage{amsmath}
\usepackage{amsthm}
\usepackage{amsfonts}
\usepackage{color}
\usepackage{enumitem}
\usepackage{xspace}
\usepackage{booktabs,colortbl}
\usepackage{array}

\newcommand\emb{\mathrm{emb}}
\newcommand\Emb{\mathrm{Emb}}
\newcommand\Att{\mathrm{Att}}
\newcommand\ctx{\mathrm{ctx}}
\newcommand\kn{\mathrm{kn}}
\newcommand\similarity{\mathrm{sim}}
\newcommand\LSTM{\mathrm{LSTM}}
\newcommand\BiLSTM{\mathrm{BiLSTM}}
\newcommand\tfidf{\mathrm{tfidf}}

\newcommand\qcSet{\ensuremath{\mathcal{S}}\xspace}
\newcommand\tok[2]{\ensuremath{w_{#1}^{#2}}\xspace}
\newcommand\e[2]{\ensuremath{e_{#1}^{#2}}\xspace}

\DeclareMathOperator*{\argmax}{arg\,max}
\DeclareMathOperator*{\argmin}{arg\,min}
\DeclareMathOperator*{\softmax}{softmax}

\newcommand\T{\rule{0pt}{2.6ex}}       

\newcommand\obqa{OpenBookQA\xspace}
\newcommand\dataurl{\url{http://data.allenai.org/OpenBookQA}\xspace}

\newcommand\fullsize{5957\xspace}
\newcommand\trainsize{4957\xspace}
\newcommand\devsize{500\xspace}
\newcommand\testsize{500\xspace}
\newcommand\factsize{1326\xspace}
\newcommand\trainname{Train\xspace}
\newcommand\devname{Dev\xspace}
\newcommand\testname{Test\xspace}

\newcommand\mcq{q_{\mathrm{mc}}\xspace}

\newcommand\cs{\ensuremath{\{c_1, c_2, c_3, c_4\}}}
\newcommand\ci{\ensuremath{c_i}\xspace}

\newcommand\Fs{\ensuremath{\mathcal{F}}\xspace}
\newcommand\Qs{\ensuremath{\mathcal{Q}}\xspace}
\newcommand\Ks{\ensuremath{\mathcal{K}}\xspace}

\usepackage[textsize=tiny, textwidth=2.5cm]{todonotes}
\usepackage{marginnote}
\newcounter{todonumber}
\newcommand{\note}[2][]{{%
 \let\marginpar\marginnote%
 \ifodd\value{todonumber}%
   \reversemarginpar%
 \else%
 \fi%
 \todo[#1]{#2}}%
 \stepcounter{todonumber}%
}

\newcommand{\revtwo}[1]{#1}
\newcommand{\incorrect}[1]{\textcolor{red}{#1}}

\usepackage{color,soul}
\setul{0.5ex}{0.4ex}

\newcommand{\tc}[2]{\setulcolor{#1}\ul{#2}\setulcolor{black}}

\newcommand{\tcA}[1]{\tc{blue}{#1}}
\newcommand{\tcB}[1]{\tc{orange}{#1}}

\newcommand{\tcD}[1]{\tc{green}{#1}}
\newcommand{\tcE}[1]{\tc{brown}{#1}}
\newcommand{\tcF}[1]{\tc{cyan}{#1}}
\newcommand{\tcG}[1]{\tc{lightgray}{#1}}
\newcommand{\tcH}[1]{\tc{darkgray}{#1}}

\newcommand{\tcJ}[1]{\tc{violet}{#1}}

\newcommand{\tcK}[1]{\tc{gray}{#1}}
\newcommand{\tcL}[1]{\tc{olive}{#1}}
\newcommand{\tcM}[1]{\tc{magenta}{#1}}

\title{Can a Suit of Armor Conduct Electricity?\\
A New Dataset for Open Book Question Answering
}

\author{
Todor Mihaylov$^\ddagger$ \and Peter Clark$^\dagger$ \and Tushar Khot$^\dagger$ \and Ashish Sabharwal$^\dagger$\\
\\
$^\dagger$ Allen Institute for Artificial Intelligence, Seattle, WA, U.S.A.\\
$^\ddagger$ Research Training Group AIPHES \& Heidelberg University, Heidelberg, Germany\\
{\tt \small \{peterc,tushark,ashishs\}@allenai.org, mihaylov@cl.uni-heidelberg.de}
}

\date{}

\begin{document}

\maketitle

\begin{abstract}
We present a new kind of question answering dataset, \obqa, modeled after open book exams for assessing human understanding of a subject. The open book that comes with our questions is a set of \factsize elementary level science facts. Roughly 6000 questions probe an understanding of these facts and their application to novel situations. This requires combining an open book fact (e.g., metals conduct electricity) with broad common knowledge (e.g., a suit of armor is made of metal) obtained from other sources. While existing QA datasets over documents or knowledge bases, being generally self-contained, focus on linguistic understanding, \obqa probes a deeper understanding of both the topic---in the context of common knowledge---and the language it is expressed in. Human performance on \obqa is close to 92\%, but many state-of-the-art pre-trained QA methods perform surprisingly poorly, worse than several simple neural baselines we develop. Our oracle experiments designed to circumvent the knowledge retrieval bottleneck demonstrate the value of both the open book and additional facts. We leave it as a challenge to solve the retrieval problem in this multi-hop setting and to close the large gap to human performance.
\end{abstract}

\section{Introduction}
\label{sec:intro}

\begin{figure}[t]
\centering
\fbox{
\begin{minipage}{0.44\textwidth}
\small
\textbf{Question:}\\
\emph{Which of these would let the most \tcA{heat} \tcB{travel} through?}\\
A) a \tcK{new} pair of \tcL{jeans}. \\
B) a \tcM{steel} \tcE{spoon} in a cafeteria. \\
C) a \tcF{cotton} \tcG{candy} at a store. \\
D) a calvin klein \tcF{cotton} \tcH{hat}. \\

\noindent
\textbf{Science Fact:} \\
\tcD{Metal} is a \tcJ{thermal conductor}. \\

\noindent
\textbf{Common Knowledge}:\\
\tcM{Steel} is made of \tcD{metal}.\\
\tcA{Heat} \tcB{travels} through a \tcJ{thermal conductor}.
\end{minipage}
}  
\caption{\label{figure:dataset-example}  An example for a question with a given set of choices and supporting facts.}
\end{figure}

Open book exams are a common mechanism for assessing human understanding of a subject, where test takers are allowed free access to a relevant book, study guide, or class notes when answering questions. In this context, the goal is not to evaluate memorization but a deeper understanding of the material and its application to new situations~\cite{Jenkins1995OpenBA,Landsberger1996StudyGS}. The application, in turn, often requires combining a fact in the book (e.g., \emph{metals conduct electricity}) with additional common knowledge the test taker is expected to have acquired by this stage (e.g., \emph{a suit of armor is made of metal}).

Motivated by this setting, we present a new kind of question answering dataset, \obqa,\footnote{The dataset and the code for the models are available at \dataurl.} that consists of two parts: \Qs, a set of \fullsize multiple-choice questions, and \Fs, a set of \factsize diverse facts about elementary level science. $\mathcal{F}$ has three key characteristics of an `open book': (a) it forms the basis for generating \Qs; (b) it has been deemed central to scientific explanations~\cite{Jansen2018WorldTreeAC}; and (c) by itself, $\mathcal{F}$ is generally insufficient to answer questions in \Qs. Faced with a question $q \in \Qs$, a student or system $S$ is expected retrieve a relevant fact $f \in \Fs$, and appeal to their own common knowledge, $\mathcal{K_S}$, when applying $f$ to answer $q$.

Figure~\ref{figure:dataset-example} provides an example. Here, \emph{metals are thermal conductors} is a core scientific fact available in \Fs. One way to apply this fact to decide whether \emph{a steel spoon} would let the \emph{most heat travel through} is to appeal to common knowledge that steel is metallic and heat travels through thermal conductors. In general, the expected common knowledge is relatively simple (taxonomic facts, definitions, object properties, etc.); the difficulty lies in identifying it and meaningfully combining it with a core fact from \Fs to answer the question.

\obqa questions are challenging as they require \emph{multi-hop reasoning with partial context} provided by $\mathcal{F}$. Specifically, unlike existing datasets for reading comprehension (RC), answering questions on the back of a textbook (TQA),\footnote{Only $\sim$5\% of the TQA questions of \citet{Kembhavi2017AreYS} require additional common knowledge.} as well as question answering over structured knowledge-bases (KBQA), the open book $\mathcal{F}$ that comes with \obqa is not self-contained. A successful system must therefore go beyond the typical challenges such as \revtwo{paraphrase matching and} coreference resolution, without benefiting from the canonicalized and complete information in KBQA.

Generating interesting open book questions is a difficult task. We used a multi-stage process starting with $\mathcal{F}$, using crowd-sourcing to generate (noisy) questions based on $\mathcal{F}$ that probe novel situations, using an automatic filter to ensure hardness for retrieval and association based systems, using a crowd filter to ensure answerability by a lay person, and further using an expert filter to ensure higher quality in \devname and \testname sets.

We evaluate a number of existing QA systems for science (without retraining) on \obqa, finding that they perform surprisingly close to the random guessing baseline of 25\%. Human performance, on the other hand, is close to 92\%.\footnote{To avoid ambiguity in the term `human performance', Section~\ref{subsec:human} describes the specific randomized model we use.}

Motivated by recent findings of gameability of NLP datasets~\cite{Gururangan2018AnnotationAI}, we also develop and evaluate simple, attention-based, neural baselines including a \emph{plausible answer detector} (which ignores the question text completely) and an \emph{odd-one-out solver}. These highlight inevitable human bias in any crowdsourced dataset, increasing performance on \obqa to 48\%.

Building upon a recent neural model for incorporating external knowledge in the story cloze setting~\cite{Mihaylov2018EnhanceCS}, we propose a knowledge-aware neural baseline that can utilize both the open book $\mathcal{F}$ and common knowledge retrieved from sources such as ConceptNet~\cite{Speer2017Conceptnet55}. While retrieving the most useful pieces of knowledge remains an open challenge, our `oracle' experiments with the fact 
$f$ 
used while generating a question 
$q$ 
and an interpretation (by the question author) of the additional knowledge 
$k$ 
needed for 
$q$, 
provides valuable insight into the nature of this dataset: Facts from the open book \Fs are valuable (5\% improvement) but not sufficient. Using both $f$ and $k$ increases the accuracy to 76\%, but is still far from human level performance, suggesting the need for non-trivial reasoning to combine these facts.

To encourage further research on this new task,
for each \trainname and \devname question $q$, \obqa also includes $f$ as intermediate supervision signal, which may be viewed as a partial \emph{explanation} for $q$.
We leave closing the large gap to human performance as a challenge for the NLP community.


\section{Related Work}
\label{sec:related}


By construction, answering \obqa questions requires
(i) some base science facts from a provided `open book', (ii) broader understanding about the world (common or commonsense knowledge), and (iii) an ability to combine these facts (reasoning). This setup differs from several existing QA tasks, as summarized below.

Reading Comprehension (RC) datasets have been proposed as benchmarks to evaluate the ability of systems to understand a document by answering factoid-style questions over this document. These datasets have taken various forms: multiple-choice~\cite{Richardson2013-mctest-dataset}, cloze-style~\cite{Hermann2015-rc-cnn-dm,Onishi2016-rc-whodidwhat,Hill2016-booktest}, and span prediction~\cite{Rajpurkar2016-squad,Trischler2017-rc-newsqa,joshi-EtAl:2017:Trivia-qa}
However, analysis \cite{Chen2016-stanford-reader,Sugawara2016-rc-skills} of these datasets has shown that many of the questions can be solved with context token matching \cite{Chen2016-reading-wikipedia-qa,Weissenborn2016-FastQa} or relatively simple paraphrasing.

To focus on the more challenging problem of reasoning across sentences, new datasets have been proposed for multi-step RC.
%
\textbf{QAngaroo}~\citep{QAngaroo:Welbl2017ConstructingDF} have used a knowledge-base to identify entity pairs (s, o) with a known relation, r, which is also supported by a multi-hop path in a set of documents. They use structured tuple queries (s, r, ?) and use all the documents along the path as the input passage.    
\textbf{NarrativeQA}~\citep{NarrativeQADeepMind2017} is an RC dataset that has been shown to require an iterative reasoning about the narrative of a story. Similar to \obqa, the questions were generated to ensure that the answer is not a direct match or paraphrase that can be retrieved with an IR approach. 
Most recently, \citet{MultiRCKhashabi2018} proposed \textbf{MultiRC}, a multiple-choice RC dataset that is designed to require multi-sentence reasoning and can have multiple correct answers. Again, like most RC datasets, it is self-contained.

\paragraph{Tasks with external knowledge.} While many of the RC datasets could benefit from commonsense or background knowledge, they are designed to be self-contained, i.e., solvable by the document context alone. Datasets such as the \textbf{Story Cloze Test}~\cite{Mostafazadeh2016AStories}, \textbf{MCScript},\footnote{SemEval-2018 Task 11: Machine Comprehension using Commonsense Knowledge \url{https://competitions.codalab.org/competitions/17184}} and \textbf{ProPara}~\cite{Propara:Mishra2018TrackingSC} do require additional domain knowledge about everyday events, scripts, and processes, respectively. However, these datasets need domain-specific modeling of events, whereas \obqa appeals to broad common knowledge cutting across a variety of types and topics.

\citet{Stasaski2017MultipleCQ} explore the creation of multi-hop questions and propose generating stronger distractors for the multiple-choice setting. Their work, however, starts with structured knowledge, specifically a Biology ontology.

Lastly, many \textbf{Science Question Answering} datasets~\cite[e.g.][]{clark2016combining,ARCClark2018} have been released that need broad external knowledge to answer the questions. However, these questions are not associated with a core set of facts, i.e., an  ``open book'' used to define these questions. As a result, the questions vary widely in style and complexity~\cite{ARCClark2018}.
In contrast, \obqa focuses on a more well-defined subset of science QA, appealing to one core fact from the open book and one (or few) relatively simple commonly known supporting facts.


\section{\obqa Dataset}
\label{sec:dataset}

\revtwo{The \obqa dataset consists of about 6,000 4-way multiple-choice questions, each associated with one core fact from a ``book'' \Fs of \factsize such facts, and an auxiliary set \Ks of about 6000 additional facts. The questions were} created via a multi-stage crowdsourcing and partial expert filtering process, discussed in Section~\ref{subsec:process}.

\revtwo{The small ``book'' \Fs consists of recurring science themes and principles, each of which can be (and here is) instantiated into multiple questions. For \Fs, we use a subset of the WorldTree corpus which \citet{Jansen2018WorldTreeAC} have analyzed for sufficiency for elementary level science. The subset we use is taken from the 
2287
WorldTree facts that were marked as ``central'' by the original authors in at least one explanation. We further filter them down to \factsize that appear general enough to be applicable to multiple situations.}

\revtwo{\obqa additionally requires broad common knowledge, which is expected to come from large corpora, such as ConceptNet, Wikipedia, or a corpus with 14M science-related sentences used by some existing baselines. The crowdsourcing process below also asks workers to mark a second fact, $k$, needed for each question $q$, in addition to $f$. These second facts, unfortunately, were often incomplete, over-complete, or only distantly related to $q$. We thus include in \obqa the set \Ks of such second facts only as \emph{auxiliary data} for optional use. We emphasize that \Ks should not be viewed as `gold' additional facts, or as a substitute for broad common knowledge.}

\begin{figure*}[t]
\centering
\includegraphics[scale=0.67]{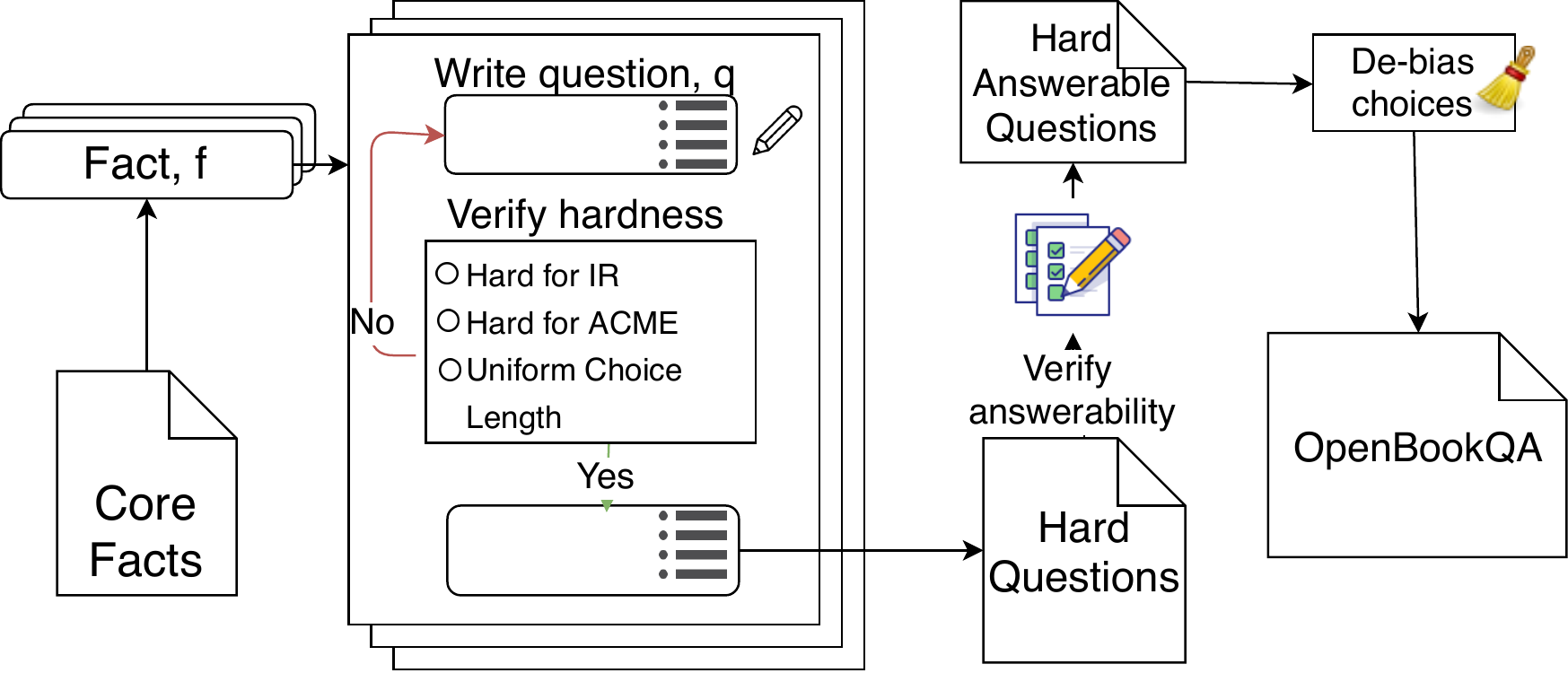}
\caption{\label{fig:generation} \obqa question generation pipeline}
\end{figure*}

\subsection{Crowdsourcing Process}
\label{subsec:process}

The overall question generation and filtering pipeline is summarized in Figure~\ref{fig:generation}. Given the ``book'' \Fs of core facts, the process proceeds as follows, starting with an empty question set $Qs$ and an empty `second facts' set \Ks:
\begin{enumerate}[wide, labelwidth=!, labelindent=4pt, itemsep=0pt]

\item A crowd-worker\footnote{\label{footnote:amt} We used Amazon Mechnical Turk, with workers from North America and with a `masters' level qualification.} $w$ is shown a random science fact $f$ from the set \Fs.

\item $w$ is asked to think of a second common fact, $k$, that may be combined with $f$ to derive a new, valid assertion $s$.

\item \label{step:creation} $w$ then converts $s$ into a question-answer pair and extends this into a 4-way multiple choice question by adding 3 incorrect answer choices, $\mcq = (q, \cs)$, where one of the \ci's is the unique correct answer.

\item \label{step:basic-checks} The system verifies $\mcq$ passes basic checks such as uniformity of answer choices.\footnote{Specifically, it looks for: 1) exactly 4 answer choices; 2) no negation words to trivially fool baselines (no, none, not, isn't, doesn't, aren't, don't, won't, except, can't, shouldn't, wouldn't, couldn't, mustn't);
3) uniform answer choice length: all with at most 3 or at least 4 words.}

\item $w$ then feeds the multiple-choice question $\mcq$ to an information retrieval solver~\cite{clark2016combining} and a word association based solver \cite{Turney2017LeveragingTB}, and verifies that (a) neither of them answers $\mcq$ correctly and (b) the top 3 IR retrieved sentences are insufficient to answer $\mcq$; if not, the question is edited and re-tried.

\item \label{step:answerability} Question $\mcq$ is then shown to 5 new crowd-workers, who are asked to answer it.

\item If at least 4 out of 5 workers answer $\mcq$ correctly, it is deemed answerable and the process continues. If not, $\mcq$ is discarded.

\item The answer choices of $\mcq$ are randomly shuffled to avoid unintended bias.\footnote{Choice `A' was the correct answer in 69\%  of the questions at the end of Step~\ref{step:basic-checks}.
}

\item $\mcq$ is associated with $f$ as the core science fact
and added to the question set \Qs. \revtwo{$k$ is added to the set \Ks of additional (noisy) facts.}

\revtwo{The \devname and \testname splits were further filtered by an in-house expert to ensure higher quality.}

\end{enumerate}

\subsection{Human Performance}
\label{subsec:human}

To assess human accuracy on this dataset, we consider the following model: Each question $q \in \Qs$ has some (unknown) human accuracy $p_q$, defined as the probability that a random human subject, chosen uniformly from a large pool $\mathcal{H}$, would answer $q$ correctly. Thus, we can think of this as defining a Bernoulli random variable, $X_q \sim B(p_q)$, whose mean is (unknown) $p_q$. The average human accuracy on $\Qs$ under this model is:
\[
  H(\Qs) = \frac{1}{|\Qs|} \, \sum_{q \in \Qs} p_q
\]
where $\{p_q \mid q \in \Qs\}$ are unknown.

With $\mathcal{H}$ as the set of crowd-workers (cf.\ Footnote~\ref{footnote:amt}), step~\ref{step:answerability} of the above question generation process is equivalent to obtaining 5 independent samples, $X_{q,i}, i \in I, |I|=5$, from $B(p_q)$. We must, however, be careful when using this data to estimate $p_q$, as the same 5 samples were used to decide whether $q$ makes it into the question set \Qs or not. For instance, if we had kept only those questions that all 5 workers answered correctly, it would clearly be inaccurate to claim that the human accuracy on $\Qs$ is 100\%. Nevertheless, it is possible to re-use the judgments from Step~\ref{step:answerability} to approximate $H(\Qs)$ with high confidence, without posing the questions to new workers.

Intuitively, if all questions in $\Qs$ were difficult to answer (i.e., all $p_q$ were small), it would be unlikely that all $|\Qs|$ questions would pass the test in Step~\ref{step:answerability}. We can use the contrapositive of this observation to conclude that $p_q$, on average, must have been high for $q \in \Qs$.

Formally, aggregating across all questions gives the following empirical estimate of $H(\Qs)$:
\begin{align*}
  \tilde{H}(\Qs) & = \frac{1}{|\Qs|} \, \sum_{q \in \Qs} \, \frac{1}{|I|} \, \sum_{i \in I} X_{q,i}\\
    & = \frac{1}{|\Qs| |I|} \, \sum_{q \in \Qs, i \in |I|} \, X_{q,i}
\end{align*}

For analysis, we assume all samples $X_{q,i}$ are independent, i.e., every answer is obtained independently.\footnote{Realistically, there is some dependence across questions as a single worker may answer multiple questions. We leave a formal analysis of this setting as future work.} An application of Hoeffding's Inequality~\cite{hoeffding1963probability} shows that $\tilde{H}(\Qs)$ converges to $H(\Qs)$ very rapidly as $n = |\Qs| |I|$ grows; specifically, $\tilde{H}(\Qs) \leq H(\Qs) + t$ with probability at least $1 - \exp(-2 n t^2)$; similarly for $\tilde{H}(\Qs) \geq H(\Qs) - t$. In our Dev and Test sets, where $|\Qs| = 500$ and $|I| = 5$, this translates into $H(Q)$ being at least $\tilde{H}(\Qs) - 3\%$ with probability over 98.8\% and at least $\tilde{H}(\Qs)- 2.5\%$ with prob 95.6\%; we report the former as our conservative estimate on human performance.


\newcommand{\property}{\textsc{Property}\xspace}
\newcommand{\causal}{\textsc{Causal}\xspace}
\newcommand{\isa}{\textsc{Isa}\xspace}
\newcommand{\basic}{\textsc{Basic}\xspace}
\newcommand{\defn}{\textsc{Definition}\xspace}
\newcommand{\common}{\textsc{Common-Sense}\xspace}
\newcommand{\others}{\textsc{Others}\xspace}

\newcommand{\question}[1]{\textcolor{Violet}{#1}}
\newcommand{\correct}[1]{\textcolor{OliveGreen}{\textbf{#1}}}

\subsection{Question Set Analysis}
\label{subsec:analysis}

\obqa consists of \fullsize questions, with \trainsize/\devsize/\testsize in the \trainname/\devname/\testname splits.\footnote{\revtwo{Overall, 8140 questions were collected, of which 2183 were discarded in crowdsourcing Step 7.}}
Table \ref{table:statistics} summarizes some statistics about the full dataset. Each question has exactly four answer choices and one associated fact used in the creation process.
We report the average length of questions, candidate choices, and associated facts, as well as how often is the longest/shortest choice the correct one.

\begin{table}[t]
\centering
\setlength\doublerulesep{\arrayrulewidth}
\setlength\tabcolsep{7pt}
\begin{tabular}{lr}
\hline
\multicolumn{2}{c}{\obqa Statistics} \\
\hline \hline
\# of questions               & 5957          \\
\# of choices per question    & 4             \\
Avg. question sentences       & 1.08 (6)      \\
Avg.~question tokens          & 11.46 (76)    \\
Avg.~choice tokens            & 2.89 (23)     \\
Avg.~science fact tokens      & 9.38 (28)     \\
Vocabulary size (q+c)         & 11855         \\
Vocabulary size (q+c+f)       & 12839         \\
Answer is the longest choice  & 1108 (18.6\%) \\
Answer is the shortest choice & 216 (3.6\%)   \\
\hline
\end{tabular}
\caption{Statistics for full \obqa dataset. Parenthetical numbers next to each average are the \textit{max}.}
\label{table:statistics}
\end{table}

We analyzed 100 questions in the \trainname set to capture the kind of common knowledge and reasoning needed. For each, we wrote down the additional common knowledge needed to answer this question \emph{in addition} to the original science fact. In 21\% of the cases, the crowdsourced question actually tests for a fact that doesn't necessarily need the original science fact. For example, the question: ``On a rainy day the clouds are (A) low (B) white (C) small (D) gray'' was written based on the science fact ``clouds produce rain'' but doesn't need this fact to answer it. We ignore such questions in our analysis. For the remaining questions, we categorized the additional facts into five high-level categories (and collapsed the remaining facts into a catch-all \others category) based on previous approaches on similar science questions~\cite{ARCClark2018,Jansen2016WhatsIA}:

\begin{enumerate}[itemsep=-2pt, itemindent=-4pt]
\item 
\isa: Basic taxonomic facts such as isa(tree, living thing), isa(granite, rock).

\item 
\property: Properties of objects such as madeof(belt buckle, metal), has(mammals, four legs), contains(lemon juice, citric acid). 

\item 
\defn: Definitions of objects that may be based on their appearance (tape is a plastic with markings), working mechanism (telescope is a device that uses mirrors to view objects), etc. 

\item 
\causal: Causal facts such as causes(adding lemon juice to milk, milk to break down). 

\item 
\basic: General scientific fact that did not fit above, e.g. squirrels eat nuts for food.  
\end{enumerate}

\begin{table}[t!]
\centering
\setlength\doublerulesep{\arrayrulewidth}
\begin{tabular}{lrr}
\hline
Fact Type & \% Questions & \% Facts \\
\hline\hline
\property & 29.11\% & 25.81\% \\
\isa      & 20.25\% & 17.20\% \\
\basic    &  17.72\% & 19.35\% \\
\defn     & 17.72\% & 15.05\% \\
\causal   &  11.39\% &  9.68\% \\
\others  & 13.92\% & 12.90\% \\
\hline
\end{tabular}
\caption{\label{table:dataset-percentages} Percentage of questions and facts for the five most common type of additional facts. Note that \% Questions does not add up to 100\% since we count the percentage of questions where at least one such fact is needed.
}
\end{table}

Table~\ref{table:dataset-percentages} presents the proportions of these facts in our analyzed question set. For each type of fact, we calculate the percentage of questions that need at least one such fact (shown as \% Questions). We also calculate the overall percentage of each fact type across all the common knowledge facts (shown as \% Facts). Most of our questions need simple facts such as $isa$ knowledge and properties of objects, further confirming the need for simple 
\revtwo{reasoning with common knowledge.}
Apart from these five major categories of facts, the catch-all \others category contains common-sense facts (e.g., it is dark at night), world knowledge (e.g., Japan is often hit by earthquakes) and lexical rewrites\footnote{Of course, every question had lexical variations. We marked it when this was the \emph{only} change to the core fact.} (e.g., \emph{ad infinitum} means over and over). 

Most of our questions need simple facts that should be easily retrievable from any knowledge-base/textual corpora. On an average, each question needed 1.16 additional facts ignoring any linguistic variations. Despite the simplicity of the knowledge needed for these questions, as we show empirically, most baseline approaches achieve a relatively low score on this dataset (even when the core fact is provided). We claim that this is due to the fact that the reasoning needed to answer these questions is non-trivial. Table~\ref{table:dataset-analysis} shows few questions with the associated facts and high-level reasoning needed to answer these questions. Assuming a model can extract the described relations (e.g. defn, contains), the QA system still needs to be able to chain these facts together, identify the resulting relation and verify its expression for each choice. In the extreme case (as shown in the last example), even though only one additional fact is needed to answer the question, it needs a system to \emph{apply} the core ``general'' science fact to a ``specific'' situation.

\begin{table*}
\centering
\begin{small}
\setlength\doublerulesep{\arrayrulewidth}
\setlength\tabcolsep{6pt}
\begin{tabular}{>{\raggedright\arraybackslash}p{6.5cm}>{\raggedright\arraybackslash}p{2.5cm}>{\raggedright\arraybackslash}p{3.0cm}>{\raggedright\arraybackslash}p{2.2cm}}
\hline
Question & Science Fact	& Common Knowledge (Type) & Reasoning Challenge \\
\hline \hline
\T \question{What is the most likely to be an effect of acid rain on an aquatic environment?} (A) increase in plant growth (B) increase in fish population \correct{(C) decrease in plant life} (D) cleaner and clearer water & acid rain has a negative impact on water quality & decrease in water quality leads to a decrease in aquatic life (\causal) & causes(x, y) $\land$ causes(y, z) $\Rightarrow$ causes(x, z) \\
\hline
\T \question{The moon's surface} (A) is smooth on the entire surface (B) contains an internal core of cheese (C) is filled with lakes \correct{(D) contains large cavities cause by explosions} & the moon's surface contains many craters & Craters are large cavities caused by explosions (\defn) & contains(x, y) $\land$ defn(y, z) $\Rightarrow$ contains(x, z) \\
\hline
\T \question{As a car approaches you in the night} (A) the headlights remain at a constant (B) the headlights turn off \correct{(C) the headlights become more intense} (D) the headlights recede into the dark & as a source of light becomes closer, that source will appear brighter & Headlights of a car are source of light (\property) & [lhs $\Rightarrow$ rhs] $\Rightarrow$ [ground(lhs) $\Rightarrow$ \hspace*{1ex}~ground(rhs)] \\
\hline
\end{tabular}
\caption{\label{table:dataset-analysis} \revtwo{Example} training \question{questions} (with their correct choices \correct{marked}) along with the facts and reasoning needed. In the last example, the science fact states that lhs=``source of light becomes closer'' implies  rhs=``source will appear brighter''. Grounding this rule based on the common-knowledge fact, produces a new rule: ``As headlights of the car come closer, headlights will appear brighter''}
\end{small}
\end{table*}

\section{Baseline Models}
\label{sec:models}

We evaluate the performance of several baselines systems on the \devname and \testname subsets of \obqa.
For each question, a solver receives 1 point towards this score if it chooses the correct answer, and $1/k$ if it reports a $k$-way tie that includes the correct answer. The ``Guess All'' baseline, which always outputs a 4-way tie, thus achieves a score of 25\%, same as the expected performance of a uniform random baseline.

\subsection{No Training, External Knowledge Only}
\label{subsec:external-only}

Since \obqa is a set of elementary level science questions, one natural baseline category is existing systems that have proven to be effective on elementary- and middle-school level science exams. These pre-trained systems, however, rely only on their background knowledge and do not take the set $\mathcal{F}$ of core facts into account. Further, their knowledge sources and retrieval mechanism are close to those used by the IR solver that, by design, is guaranteed to fail on \obqa. These two aspects place a natural limit on the effectiveness of these solvers on \obqa, despite their excellent fit for the domain of multiple-choice science questions.  We consider four such solvers. 

\textbf{PMI}~\cite{clark2016combining} \revtwo{uses pointwise mutual information (PMI) to score each answer choice using statistics based on a corpus of 280 GB of plain text. It extracts unigrams, bigrams, trigrams, and skip-bigrams from the question $q$ and each answer choice $c_i$. Each answer choice is scored based on the average PMI across all pairs of question and answer n-grams.}

\textbf{TableILP}~\cite{tableilp2016} is an \revtwo{Integer Linear Programming (ILP) based reasoning system designed for science questions. It operates over semi-structured relational tables of knowledge. It scores each answer choice based on the optimal (as defined by the ILP objective) ``support graph'' connecting the question to that answer through table rows.} The small set of these knowledge tables, however, often results in missing knowledge, making TableILP not answer 24\% of the \obqa questions at all.

\textbf{TupleInference}~\cite{Khot2017AnsweringCQ}, \revtwo{also an ILP-based QA system, uses Open IE tuples~\cite{Banko2007OpenIE} as its semi-structured representation. It builds these subject-verb-object tuples \emph{on-the-fly} by retrieving text for each question from a large corpus. It then defines an ILP program to combine evidence from multiple tuples.}

\textbf{DGEM}~\cite{Khot2018Scitail} is a neural entailment model that also uses Open IE to produce a semi-structured representation. \revtwo{We use the adaptation of this model to multiple-choice question answering proposed by \citet{ARCClark2018}, which works as follows:}
(1) convert $q$ and each \ci into a hypothesis, $h_i$, and each retrieved fact into a premise $p_j$; and (2) return the answer choice with the highest entailment score, $\argmax_i e(p_j, h_i)$.

\subsection{No Training; $\mathcal{F}$ and Extr.~Knowledge}
\label{subsec:existing-with-core-facts}

We also consider providing the set $\mathcal{F}$ of core facts to two existing solvers: the \textbf{IR} solver of \citet{clark2016combining} (to assess how far simple word-overlap can get), and the \textbf{TupleInference} solver.

\subsection{Trained Models, No Knowledge}
\label{subsec:neural-no-external}

We consider several neural baseline models that are trained using \trainname set of \obqa. For ease of explanation, we first define the notation used in our models. For a given question $\mcq = (q, \cs)$, we define the set of 
token sequences
, $\qcSet = \{q, c_1, c_2, c_3, c_4\}$. For each token sequence $s \in \qcSet$, $\tok{j}{s}$ is the $j^{th}$ 
and $\e{j}{s} = \Emb(\tok{j}{s})$ is the embedding for this token. We use $n_s$ to indicate the number of tokens in $s$ and $d$ for the dimensionality of the embeddings.\footnote{For all experiments we use $d=300$ \emph{GloVe}~\cite{Pennington2014-glove} embeddings pre-trained on 840B tokens from \textit{Common Crawl} (https://nlp.stanford.edu/projects/glove/).} We model multiple-choice QA as multi-class classification: Given $\mcq$, predict one of four class labels $L=\{1,2,3,4\}$, where the true label is the correct answer index. 

\paragraph{Embeddings + Similarities as Features.} We first experiment with a simple logistic regression model
\cite{SemEval2016:task3:SemanticZ,mihaylovfrank:2016,mihaylovfrank:2017}
that uses centroid vectors $r^{\emb}_s$ of the word embeddings of tokens in $s$, and then computes the cosine similarities between the question and each answer choice, $r^{\cos}_{q, \ci}$:
\begin{align*}
r^{\emb}_{s} & = \frac{1}{n_s} \sum_{j=1}^{n_s} e_{s_j} \in \mathbb{R}^d\\
r^{\cos}_{q,c_{i}} &= \cos(r^{\emb}_{q}, r^{\emb}_{c_i}) \in \mathbb{R}^1
\end{align*}
For each training instance, we build a feature representations $\vec{f}$ by concatenating these vectors and train an $L2$ logistic regression classifier:
\begin{align*}
\vec{f} = [r^{\emb}_{q} ; r^{\emb}_{c_{1..4}} ; r^{\cos}_{q,c_{1..4}}] \in \mathbb{R}^{5d+4}
\end{align*}
\paragraph{BiLSTM Max-Out Baselines.} As a simple neural baseline, we adapt \textit{BiLSTM max-out} model~\citep{Conneau2017} to our QA task. That is, we first encode the question tokens and choice tokens $\tok{1\ldots n_s}{s}$, independently with a bi-directional context encoder ($\LSTM$) to obtain a context ($\ctx$) representation 
$h^{\ctx}_{s_{1\ldots n_s}} = \BiLSTM(\e{1\ldots n_s}{s}) \in \mathbb{R}^{n_s \times 2h}$
Next, we perform an element-wise aggregation operation $\max$ on the encoded representations $h^{\ctx}_{s_{1..n_s}}$ to construct a single vector:
\begin{align}
\label{eq:max-h}
r^{\ctx}_{s} = \max(h^{\ctx}_{s_{1..n_s}}) \in \mathbb{R}^{2h}.
\end{align}

Given the contextual representations for each token sequence, we experiment with three configurations for using these representations for QA:

\paragraph{(a) Plausible Answer Detector.} This baseline goes to the extreme of completely ignoring $q$ and trying to learn how plausible it is for $c_i$ to be the correct answer to \emph {some} question in this domain. This captures the fact that certain choices like `a magical place' or `flying cats' are highly unlikely to be the correct answer to a science question without negation (which is the case for \obqa).

We implement a plausible answer detector using a \textit{choice-only} model for predicting the answer by obtaining a score $\alpha_{c_i}$ as:
$\alpha_{c_i} = W^T_{c}r^{\ctx}_{c_i} \in \mathbb{R}^{1},$
where $ W^T_{c}  \in \mathbb{R}^{2h}$ is a weights vector optimized during training, $i=\{1..4\}$ is the index of the choice.
To obtain the answer choice 
from the set of choice scores $\alpha_{c_{1..4}}$ 
using
$\argmax(\softmax(\alpha_{c_{1..4}})),$
where
$\softmax(\alpha_{c_{i}}) = \frac{\exp(\alpha_{c_{i}})}{\sum_{j=1}^{4}{\exp(\alpha_{c_{j}})}}$ as usual. 

\paragraph{(b) Odd-One-Out Solver.} It considers all 4 answer options jointly and selects the one that is least similar to the others. This captures bias in human authored questions arising from the fact that creating good quality incorrect answers is difficult. Workers generally start with the correct answer, and then come up with three incorrect ones. The latter often tend to be homogeneous or share other common properties (e.g., non-scientific terms) uncharacteristic of the correct answer.

We implement this using a 
\textit{choice-to-choices}
attention model. For each choice $c_i$, we calculate the attention to the other choices as $\alpha_{c_i,c_j}$. We then sum these attention values to compute the attention for $c_i$ to the rest of the choices, $\alpha_{c_{i}2c_{r(est)}}$, and return the choice with the lowest sum.  The attention is computed as  
$\alpha_{c_i, c_j} = \Att(r^{\ctx}_{c_i}, r^{\ctx}_{c_j})$
where
\begin{align*}
\Att(u, v) & = W^T([u; v; u \cdot v ; |u - v|]) \in \mathbb{R}^{1}
\end{align*}
is a linear attention function 
\revtwo{and}
$W \in \mathbb{R}^{8h}$ is a weight vector. We then compute $\alpha_{c_{i}2c_{r(est)}} = \sum_{j=1}^{4}{\alpha_{c_i,c_j}}$ ($j \neq i$)
and select the answer with the index $a_{c2c_{r}} = \argmin(\softmax(\alpha_{c_{1..4}2c_{r}}))$.

\paragraph{(c) Question Match.} 
This solver tries to predict which choice best matches the question \cite{cqa-semeval-2016}, without relying on external knowledge. 
To achieve that, we compute an attention score $\alpha_{q, c_i}$ between $q$ and each of the choices $q_i$ as
$\alpha_{q, c_i} = \Att(r^{\ctx}_q, r^{\ctx}_{c_i}),$
and select the one with the highest score. We also experiment with a model where $r^{\ctx}_q$ and $r^{\ctx}_{c_i}$ are obtained using token-wise interaction proposed in ESIM~\cite{esim-chen2017}.
\subsection{Trained Model with External Knowledge}
\label{subsec:neural-with-knowledge}
Lastly, we implement a two stage model for incorporating external common knowledge, $K$. The first module performs information retrieval on $K$ to select a fixed size subset of potentially relevant facts $K_{Q, C}$ for each instance in the dataset (see Appendix A).
The second module is a neural network that takes ($Q$, $C$, $K_{Q, C}$) as input to predict the answer $a_{q, c}$ to a question $Q$ from the set of choices $C$.

\paragraph{Knowledge-Enhanced Reader.}
As a base knowledge-aware model, we use a variant of the model of \citet{Mihaylov2018EnhanceCS}, implemented by extending our \textbf{BiLSTM max-out question-match baseline (c)}. For each instance the model reads the question $q$ and answers $c_{1..4}$ independently and attends to the set of retrieved external 
knowledge facts $\mathrm{K}_{Q, C}$.
We encode each fact $k_j$ from $\mathrm{K}_{Q, C} = k_{1..N_{k}}$ ($N_{k}$ is the number of facts) with same $\BiLSTM$ 
as used for $q$ and $c_{1..4}$ and construct a single vector $r^{\ctx}_{k_j} \in \mathbb{R}^{2h}$ using Eq. \ref{eq:max-h}. 
Having such representations for each $k_j$ results in knowledge memory matrix $M_k=r^{\ctx}_{k_{1..N_{k}}} \in \mathbb{R}^{N_{k} \times 2h}$.
Note that $M_k$ is dynamic memory, specific for each instance in the batch and is encoded in each step during training. 
This memory is used to calculate a knowledge-aware representation, $r^{\kn}_{s} = \sum(({M^T_k}r^{\ctx}_{s}).{M_k}) \in \mathbb{R}^{2h}$. 
Each context ($\ctx$) representation $r^{\ctx}_{s}$ ($s \in \qcSet$) is combined with $r^{\kn}_s$ to obtain a knowledge-enhanced representation $r^{\ctx+\kn}_{s}=(r^{\ctx}_{s} + r^{\kn}_{s})/2$. 
We then model the knowledge-enhanced attention $\alpha^{\kn}_{q,c_i}$ between $q$ and $c_i$ as a linear combination of the $\ctx$, $\kn$ and $\ctx+\kn$ representations as
\begin{align*}
\alpha_{q,c_i} = W^T[\Att(r^{\ctx}_{s}, r^{\ctx}_{c_i});\Att(r^{\kn}_{s}, r^{\kn}_{c_i});\\
\Att(r^{\ctx+\kn}_{s}, r^{\ctx}_{c_i});\Att(r^{\ctx}_{s}, r^{\ctx+\kn}_{c_i});\\
\Att(r^{\ctx}_{s}, r^{\kn}_{c_i});\Att(r^{\kn}_{s}, r^{\ctx}_{c_i});\\
\Att(r^{\ctx+\kn}_{s}, r^{\kn}_{c_i});\Att(r^{\kn}_{s}, r^{\ctx+\kn}_{c_i});\\
\Att(r^{\ctx+\kn}_{s}, r^{\ctx+\kn}_{c_i})],
\end{align*}
where $W \in \mathbb{R}^{9}$ is a weight vector initialized with the $ones$ vector and optimized during training.
We then select the answer $c_i$ with the highest score.


\section{Baseline Performance}
\label{sec:results}

\begin{table}[t]
\centering
\setlength{\doublerulesep}{\arrayrulewidth}
\scalebox{0.88}{
\begin{tabular}{|lll|}
\hline
Solver & Dev & Test \\
\hline \hline
\T Human solver            & 89.3* & 91.7* \\
Guess All (``random'')  & 25.0 & 25.0 \\
\hline
\multicolumn{3}{|l|}{\T \hspace{2ex} \textsc{No Training, KB Only} (\S\ref{subsec:external-only})} \\
TupleInference          & 15.9 & 17.9 \\
PMI (Waterloo corpus)   & 19.7 & 21.2 \\ 
TableILP                & 20.0 & 23.4 \\
DGEM                    & 27.4 & {\bf 24.4} \\
\hline
\multicolumn{3}{|l|}{\T \hspace{2ex} \textsc{No Training, KB + $\mathcal{F}$} (\S\ref{subsec:existing-with-core-facts})} \\
IR with $\mathcal{F}$               & 25.5 & 24.8 \\
TupleInference with $\mathcal{F}$   & 23.6 & {\bf 26.6} \\
DGEM with $\mathcal{F}$ & 28.2 & 24.6 \\
\hline
\multicolumn{3}{|l|}{\T \hspace{2ex} \textsc{Trained Models, No $\mathcal{F}$ or KB} (\S\ref{subsec:neural-no-external})} \\
Embedd+Sim                & 44.6 & 41.8 \\ 
ESIM  & 53.9$\pm$0.4 & 48.9$\pm$1.1 \\
Plausible Answer Detector & 54.4$\pm$0.7 & 49.6$\pm$0.7 \\
Odd-one-out Solver        & 56.9$\pm$0.5 & 50.2$\pm$1.6 \\
Question Match            & 54.6$\pm$1.2 & {\bf 50.2$\pm$0.9} \\
\hline
\hline
\multicolumn{3}{|l|}{\T \hspace{2ex} \textsc{Oracle Models, $\mathcal{F}$ and/or KB} (\S\ref{subsec:neural-with-knowledge})} \\
$f$      & 63.0$\pm$2.3 & 55.8$\pm$2.3 \\
$f$ + WordNet    & 57.6$\pm$1.4 & 56.3$\pm$1.3 \\
$f$ + ConceptNet    & 57.0$\pm$1.6 & 53.7$\pm$1.5 \\
$f$ + $k$  & 80.2$\pm$1.1 & 76.9$\pm$0.7 \\\hline
\hline
\end{tabular}}
\caption{\label{table:baselines} Scores obtained by various solvers on \obqa, reported as a percentage $\pm$ the standard deviation across 5 runs with different random seeds.
Other baselines are described in the corresponding referenced section. For oracle evaluation, we use the gold science fact $f$ associated with each question, and optionally the additional fact $k$ provided by the question author. 
Bold denotes the best Test score in each category.
}
\end{table}

The results for various baseline models are summarized in Table~\ref{table:baselines}, grouped by method category. 
We make a few observations:

First, the task is \textbf{largely solvable by a lay-person}, as evidenced by the 92\% score of crowd-workers. \revtwo{This is measured as described in Section~\ref{subsec:human}. We use annotations from Step 6 of the question generation process and report $\tilde{H}(Q) - 3\%$ as a conservative lower estimate. As an additional assessment, we also obtained 5 \emph{new} annotations for 100 randomly chosen questions from each of Train, Dev, and Test sets. The performance remained similar at 88.6\%, 90.2\%, and 91.6\%, resp.}

The \textbf{second group} shows that pre-trained state-of-the-art solvers for multiple-choice science questions perform poorly. One explanation is their correlation with the the IR method used for question filtering, as mentioned in Section~\ref{subsec:external-only}.

The \textbf{third group} of results suggests that adding \Fs to pre-trained models has a mixed effect, improving TupleInference by 8.7\% but not changing DGEM.\footnote{By design, IR with its default corpus gets 0\% on \obqa. Hence we don't consider the effect of adding \Fs, which appears artificially magnified.} \revtwo{Unlike DGEM, TupleInference relies on brittle word-overlap similarity measures very similar to the ones used by IR. Since IR (KB) gets 0\% by design, TupleInference (KB) also has poor performance and adding \Fs helps it find better support despite the brittle measures.}


The \textbf{fourth group} demonstrates that carefully designed trainable neural models---even if simplistic and knowledge-free---can be surprisingly powerful. For example, the ``plausible answer detector'' can predict the correct answer with \revtwo{49.6\%} accuracy without even looking at the question. The ``odd-one-out'' solver, by considering other answer choices, raises this to \revtwo{50.2\%}. The ``question match'' solver, which simply compares the BiLSTM max-out 
\revtwo{encoding}
of the question with that of various answer choices, also achieves \revtwo{50.2}\%.\footnote{\revtwo{This model also achieves the current best score, 33.87\%, on the ARC Reasoning Challenge~\cite{ARCClark2018}.
\revtwo{When adapted for the textual entailment task by comparing BiLSTM max-out encodings of premise and hypothesis, it achieves 85\% on the SciTail dataset \cite{Khot2018Scitail}}.
}} Similar findings have been reported for several recent datasets~\cite{Gururangan2018AnnotationAI}, making it imperative to perform such tests early.

\revtwo{Interestingly, all of these neural knowledge-free baselines simultaneously succeed on 34.4\% of the Dev questions, and simultaneously fail on 23.6\%.}
For \textbf{Question Match} and \textbf{ESIM} we also experiment with ELMo \cite{elmo-Peters:2018} which improved their score on Test with 0.4\% and 1.8\%.

The \textbf{final group} demonstrates the need for external knowledge and deeper reasoning. When the ``oracle'' science fact $f$ used by the question author is provided to the knowledge-enhanced reader, it improves over the knowledge-less models by \revtwo{about 5\%}. However, there is still a large gap, showing that the core fact is insufficient to answer the question. When we also include facts retrieved from 
\revtwo{WordNet~\cite{WordNet-Miller1990},
the score improves by about 0.5\%.}
\revtwo{Unlike the WordNet gain, adding ConceptNet~\cite{Speer2017Conceptnet55} 
introduces a distraction and
reduces the score. This suggests that ConceptNet is either not a good source of knowledge for our task, or only a subset of its relations should be considered.} 
Overall, external knowledge helps, although retrieving the right bits of knowledge remains difficult.
In the last row of Table~\ref{table:baselines}, we use the oracle core fact along with question author's interpretation of the additional fact $k$. This increases the scores substantially, to about 76\%. This big jump shows that improved knowledge retrieval should help on this task. At the same time, we are still not close to the human performance level of 92\% due to various reasons: (a) the additional fact needed can be subjective, as hinted at by our earlier analysis; (b) the authored facts \Ks tend to be noisy (incomplete, over-complete, or only distantly related), also as mentioned earlier; and (b) even given the true gold facts, performing reliable ``reasoning'' to link them properly remains a challenge.

Sample predictions and analysis of questions from \devname are provided in Appendix D.

\section{Conclusion}
\label{sec:concl}

We present a new dataset, \obqa, of about 6000 questions for open book question answering. The task focuses on the challenge of combining a corpus of provided science facts (open book) with external broad common knowledge. We show that this dataset requires simple common knowledge beyond the provided core facts, as well as multi-hop reasoning combining the two. While simple neural methods are able to achieve an accuracy of about 50\%, this is still far from the human performance of 92\% on this task. We leave closing this gap for future research, and illustrate, via oracle-style experiments, the potential of better retrieval and reasoning on this task.

\vspace{-0.1cm}
\section*{Acknowledgments}
\label{sec:ack}
\vspace{-0.2cm}
\revtwo{The authors would like to thank Lane Aasen for helping develop the infrastructure for the crowdsourcing task, and Madeleine van Zuylen for providing expert annotation for the \devname and \testname questions.}

\bibliographystyle{acl_natbib_nourl}
\bibliography{open-book-qa}

\begin{thebibliography}{49}
\expandafter\ifx\csname natexlab\endcsname\relax\def\natexlab#1{#1}\fi

\bibitem[{Banko et~al.(2007)Banko, Cafarella, Soderland, Broadhead, and
  Etzioni}]{Banko2007OpenIE}
M.~Banko, M.~J. Cafarella, S.~Soderland, M.~Broadhead, and O.~Etzioni. 2007.
\newblock Open information extraction from the web.
\newblock In \emph{IJCAI}.

\bibitem[{Chen et~al.(2016)Chen, Bolton, and
  Manning}]{Chen2016-stanford-reader}
D.~Chen, J.~Bolton, and C.~D. Manning. 2016.
\newblock A thorough examination of the cnn/daily mail reading comprehension
  task.
\newblock In \emph{ACL}, pages 2358--2367.

\bibitem[{Chen et~al.(2017{\natexlab{a}})Chen, Fisch, Weston, and
  Bordes}]{Chen2016-reading-wikipedia-qa}
D.~Chen, A.~Fisch, J.~Weston, and A.~Bordes. 2017{\natexlab{a}}.
\newblock Reading wikipedia to answer open-domain questions.
\newblock In \emph{ACL}.

\bibitem[{Chen et~al.(2017{\natexlab{b}})Chen, Zhu, Ling, Wei, Jiang, and
  Inkpen}]{esim-chen2017}
Q.~Chen, X.~Zhu, Z.-H. Ling, S.~Wei, H.~Jiang, and D.~Inkpen.
  2017{\natexlab{b}}.
\newblock Enhanced lstm for natural language inference.
\newblock In \emph{ACL}, pages 1657--1668.

\bibitem[{Clark et~al.(2018)Clark, Cowhey, Etzioni, Khot, Sabharwal, Schoenick,
  and Tafjord}]{ARCClark2018}
P.~Clark, I.~Cowhey, O.~Etzioni, T.~Khot, A.~Sabharwal, C.~Schoenick, and
  O.~Tafjord. 2018.
\newblock Think you have solved question answering? {T}ry {ARC}, the {AI2}
  reasoning challenge.
\newblock \emph{CoRR}, abs/1803.05457.

\bibitem[{Clark et~al.(2016)Clark, Etzioni, Khot, Sabharwal, Tafjord, Turney,
  and Khashabi}]{clark2016combining}
P.~Clark, O.~Etzioni, T.~Khot, A.~Sabharwal, O.~Tafjord, P.~D. Turney, and
  D.~Khashabi. 2016.
\newblock Combining retrieval, statistics, and inference to answer elementary
  science questions.
\newblock In \emph{AAAI}, pages 2580--2586.

\bibitem[{Conneau et~al.(2017)Conneau, Kiela, Schwenk, Barrault, and
  Bordes}]{Conneau2017}
A.~Conneau, D.~Kiela, H.~Schwenk, L.~Barrault, and A.~Bordes. 2017.
\newblock Supervised learning of universal sentence representations from
  natural language inference data.
\newblock In \emph{EMNLP}, pages 670--680.

\bibitem[{Gardner et~al.(2017)Gardner, Grus, Neumann, Tafjord, Dasigi, Liu,
  Peters, Schmitz, and Zettlemoyer}]{Gardner2017AllenNLP}
M.~Gardner, J.~Grus, M.~Neumann, O.~Tafjord, P.~Dasigi, N.~F. Liu, M.~Peters,
  M.~Schmitz, and L.~S. Zettlemoyer. 2017.
\newblock {AllenNLP}: {A} deep semantic natural language processing platform.
\newblock \emph{CoRR}, abs/1803.07640.

\bibitem[{Gururangan et~al.(2018)Gururangan, Swayamdipta, Levy, Schwartz,
  Bowman, and Smith}]{Gururangan2018AnnotationAI}
S.~Gururangan, S.~Swayamdipta, O.~Levy, R.~Schwartz, S.~R. Bowman, and N.~A.
  Smith. 2018.
\newblock Annotation artifacts in natural language inference data.
\newblock In \emph{NAACL}.

\bibitem[{Hermann et~al.(2015)Hermann, Kocisky, Grefenstette, Espeholt, Kay,
  Suleyman, and Blunsom}]{Hermann2015-rc-cnn-dm}
K.~M. Hermann, T.~Kocisky, E.~Grefenstette, L.~Espeholt, W.~Kay, M.~Suleyman,
  and P.~Blunsom. 2015.
\newblock Teaching machines to read and comprehend.
\newblock In \emph{NIPS}, pages 1693--1701.

\bibitem[{Hill et~al.(2016)Hill, Bordes, Chopra, and
  Weston}]{Hill2016-booktest}
F.~Hill, A.~Bordes, S.~Chopra, and J.~Weston. 2016.
\newblock The goldilocks principle: Reading children's books with explicit
  memory representations.
\newblock In \emph{ICLR}.

\bibitem[{Hoeffding(1963)}]{hoeffding1963probability}
W.~Hoeffding. 1963.
\newblock Probability inequalities for sums of bounded random variables.
\newblock \emph{Journal of the American Statistical Association},
  58(301):13--30.

\bibitem[{Jansen et~al.(2016)Jansen, Balasubramanian, Surdeanu, and
  Clark}]{Jansen2016WhatsIA}
P.~Jansen, N.~Balasubramanian, M.~Surdeanu, and P.~Clark. 2016.
\newblock What's in an explanation? characterizing knowledge and inference
  requirements for elementary science exams.
\newblock In \emph{COLING}.

\bibitem[{Jansen et~al.(2018)Jansen, Wainwright, Marmorstein, and
  Morrison}]{Jansen2018WorldTreeAC}
P.~A. Jansen, E.~Wainwright, S.~Marmorstein, and C.~T. Morrison. 2018.
\newblock {WorldTree}: {A} corpus of explanation graphs for elementary science
  questions supporting multi-hop inference.
\newblock In \emph{LREC}.

\bibitem[{Jenkins(1995)}]{Jenkins1995OpenBA}
T.~Jenkins. 1995.
\newblock Open book assessment in computing degree programmes 1.
\newblock Technical Report 95.28, University of Leeds.

\bibitem[{Joshi et~al.(2017)Joshi, Choi, Weld, and
  Zettlemoyer}]{joshi-EtAl:2017:Trivia-qa}
M.~Joshi, E.~Choi, D.~Weld, and L.~Zettlemoyer. 2017.
\newblock {TriviaQA}: {A} large scale distantly supervised challenge dataset
  for reading comprehension.
\newblock In \emph{ACL}, pages 1601--1611.

\bibitem[{Kembhavi et~al.(2017)Kembhavi, Seo, Schwenk, Choi, Farhadi, and
  Hajishirzi}]{Kembhavi2017AreYS}
A.~Kembhavi, M.~J. Seo, D.~Schwenk, J.~Choi, A.~Farhadi, and H.~Hajishirzi.
  2017.
\newblock Are you smarter than a sixth grader? textbook question answering for
  multimodal machine comprehension.
\newblock In \emph{CVPR}, pages 5376--5384.

\bibitem[{Khashabi et~al.(2018)Khashabi, Chaturvedi, Roth, Upadhyay, and
  Roth}]{MultiRCKhashabi2018}
D.~Khashabi, S.~Chaturvedi, M.~Roth, S.~Upadhyay, and D.~Roth. 2018.
\newblock Looking beyond the surface: A challenge set for reading comprehension
  over multiple sentences.
\newblock In \emph{NAACL}.

\bibitem[{Khashabi et~al.(2016)Khashabi, Khot, Sabharwal, Clark, Etzioni, and
  Roth}]{tableilp2016}
D.~Khashabi, T.~Khot, A.~Sabharwal, P.~Clark, O.~Etzioni, and D.~Roth. 2016.
\newblock Question answering via integer programming over semi-structured
  knowledge.
\newblock In \emph{IJCAI}.

\bibitem[{Khot et~al.(2017)Khot, Sabharwal, and Clark}]{Khot2017AnsweringCQ}
T.~Khot, A.~Sabharwal, and P.~Clark. 2017.
\newblock Answering complex questions using open information extraction.
\newblock In \emph{ACL}.

\bibitem[{Khot et~al.(2018)Khot, Sabharwal, and Clark}]{Khot2018Scitail}
T.~Khot, A.~Sabharwal, and P.~Clark. 2018.
\newblock {SciTail}: A textual entailment dataset from science question
  answering.
\newblock In \emph{AAAI}.

\bibitem[{Kingma and Ba(2015)}]{Kingma2015-adam}
D.~P. Kingma and J.~L. Ba. 2015.
\newblock {Adam: a Method for Stochastic Optimization}.
\newblock \emph{International Conference on Learning Representations 2015},
  pages 1--15.

\bibitem[{Kocisk{\'{y}} et~al.(2017)Kocisk{\'{y}}, Schwarz, Blunsom, Dyer,
  Hermann, Melis, and Grefenstette}]{NarrativeQADeepMind2017}
T.~Kocisk{\'{y}}, J.~Schwarz, P.~Blunsom, C.~Dyer, K.~M. Hermann, G.~Melis, and
  E.~Grefenstette. 2017.
\newblock The {NarrativeQA} reading comprehension challenge.
\newblock \emph{CoRR}, abs/1712.07040.

\bibitem[{Landsberger(1996)}]{Landsberger1996StudyGS}
J.~Landsberger. 1996.
\newblock Study guides and strategies.
\newblock Http://www.studygs.net/tsttak7.htm.

\bibitem[{Mihaylov and Frank(2016)}]{mihaylovfrank:2016}
T.~Mihaylov and A.~Frank. 2016.
\newblock Discourse relation sense classification using cross-argument semantic
  similarity based on word embeddings.
\newblock In \emph{CoNLL-16 shared task}, pages 100--107.

\bibitem[{Mihaylov and Frank(2017)}]{mihaylovfrank:2017}
T.~Mihaylov and A.~Frank. 2017.
\newblock {{Story Cloze Ending Selection Baselines and Data Examination}}.
\newblock In \emph{LSDSem – Shared Task}.

\bibitem[{Mihaylov and Frank(2018)}]{Mihaylov2018EnhanceCS}
T.~Mihaylov and A.~Frank. 2018.
\newblock {Knowledgeable Reader: Enhancing Cloze-Style Reading Comprehension
  with External Commonsense Knowledge}.
\newblock In \emph{ACL}, pages 821--832.

\bibitem[{Mihaylov and Nakov(2016)}]{SemEval2016:task3:SemanticZ}
T.~Mihaylov and P.~Nakov. 2016.
\newblock {SemanticZ at SemEval-2016 Task 3}: Ranking relevant answers in
  community question answering using semantic similarity based on fine-tuned
  word embeddings.
\newblock In \emph{SemEval~'16}.

\bibitem[{Miller(1995)}]{miller1995wordnet}
G.~A. Miller. 1995.
\newblock Wordnet: a lexical database for english.
\newblock \emph{Communications of the ACM}, 38(11):39--41.

\bibitem[{Miller et~al.(1990)Miller, Beckwith, Fellbaum, Gross, and
  Miller}]{WordNet-Miller1990}
G.~A. Miller, R.~Beckwith, C.~Fellbaum, D.~Gross, and K.~J. Miller. 1990.
\newblock Introduction to {WordNet}: {A}n on-line lexical database.
\newblock \emph{International Journal of Lexicography}, 3(4):235--244.

\bibitem[{Mishra et~al.(2018)Mishra, Huang, Tandon, tau Yih, and
  Clark}]{Propara:Mishra2018TrackingSC}
B.~D. Mishra, L.~Huang, N.~Tandon, W.~tau Yih, and P.~Clark. 2018.
\newblock Tracking state changes in procedural text: A challenge dataset and
  models for process paragraph comprehension.
\newblock In \emph{NAACL}.

\bibitem[{Mostafazadeh et~al.(2016)Mostafazadeh, Chambers, He, Parikh, Batra,
  Vanderwende, Kohli, and Allen}]{Mostafazadeh2016AStories}
N.~Mostafazadeh, N.~Chambers, X.~He, D.~Parikh, D.~Batra, L.~Vanderwende,
  P.~Kohli, and J.~Allen. 2016.
\newblock {A Corpus and Evaluation Framework for Deeper Understanding of
  Commonsense Stories}.
\newblock In \emph{NAACL}.

\bibitem[{Nakov et~al.(2016)Nakov, M{\`a}rquez, Moschitti, Magdy, Mubarak,
  Freihat, Glass, and Randeree}]{cqa-semeval-2016}
P.~Nakov, L.~M{\`a}rquez, A.~Moschitti, W.~Magdy, H.~Mubarak, a.~A. Freihat,
  J.~Glass, and B.~Randeree. 2016.
\newblock Semeval-2016 task 3: Community question answering.
\newblock In \emph{SemEval~'16}, pages 525--545.

\bibitem[{Onishi et~al.(2016)Onishi, Wang, Bansal, Gimpel, and
  McAllester}]{Onishi2016-rc-whodidwhat}
T.~Onishi, H.~Wang, M.~Bansal, K.~Gimpel, and D.~McAllester. 2016.
\newblock Who did what: A large-scale person-centered cloze dataset.
\newblock In \emph{EMNLP}, pages 2230--2235, Austin, Texas.

\bibitem[{Paszke et~al.(2017)Paszke, Gross, Chintala, Chanan, Yang, DeVito,
  Lin, Desmaison, Antiga, and Lerer}]{paszke2017automatic}
A.~Paszke, S.~Gross, S.~Chintala, G.~Chanan, E.~Yang, Z.~DeVito, Z.~Lin,
  A.~Desmaison, L.~Antiga, and A.~Lerer. 2017.
\newblock Automatic differentiation in pytorch.
\newblock In \emph{NIPS-W}.

\bibitem[{Pedregosa et~al.(2011)Pedregosa, Varoquaux, Gramfort, Michel,
  Thirion, Grisel, Blondel, Prettenhofer, Weiss, Dubourg, Vanderplas, Passos,
  Cournapeau, Brucher, Perrot, and Duchesnay}]{scikit-learn}
F.~Pedregosa, G.~Varoquaux, A.~Gramfort, V.~Michel, B.~Thirion, O.~Grisel,
  M.~Blondel, P.~Prettenhofer, R.~Weiss, V.~Dubourg, J.~Vanderplas, A.~Passos,
  D.~Cournapeau, M.~Brucher, M.~Perrot, and E.~Duchesnay. 2011.
\newblock {Scikit-learn}: {M}achine learning in {P}ython.
\newblock \emph{Journal of Machine Learning Research}, 12:2825--2830.

\bibitem[{Pennington et~al.(2014)Pennington, Socher, and
  Manning}]{Pennington2014-glove}
J.~Pennington, R.~Socher, and C.~Manning. 2014.
\newblock {GloVe}: {G}lobal vectors for word representation.
\newblock In \emph{EMNLP}, pages 1532--1543.

\bibitem[{Peters et~al.(2018)Peters, Neumann, Iyyer, Gardner, Clark, Lee, and
  Zettlemoyer}]{elmo-Peters:2018}
M.~E. Peters, M.~Neumann, M.~Iyyer, M.~Gardner, C.~Clark, K.~Lee, and
  L.~Zettlemoyer. 2018.
\newblock Deep contextualized word representations.
\newblock In \emph{NAACL}.

\bibitem[{Rajpurkar et~al.(2016)Rajpurkar, Zhang, Lopyrev, and
  Liang}]{Rajpurkar2016-squad}
P.~Rajpurkar, J.~Zhang, K.~Lopyrev, and P.~Liang. 2016.
\newblock {SQuAD}: 100,000+ questions for machine comprehension of text.
\newblock In \emph{EMNLP}, pages 2383--2392.

\bibitem[{Richardson et~al.(2013)Richardson, Burges, and
  Renshaw}]{Richardson2013-mctest-dataset}
M.~Richardson, C.~J. Burges, and E.~Renshaw. 2013.
\newblock {MCTest}: {A} challenge dataset for the open-domain machine
  comprehension of text.
\newblock In \emph{EMNLP}, pages 193--203.

\bibitem[{Singh et~al.(2002)Singh, Lin, Mueller, Lim, Perkins, and
  Zhu}]{Singh2002-common-sense-kw-omcs}
P.~Singh, T.~Lin, E.~Mueller, G.~Lim, T.~Perkins, and W.~Zhu. 2002.
\newblock Open mind common sense: Knowledge acquisition from the general
  public.
\newblock In \emph{Lecture Notes in Computer Science}, volume 2519, pages
  1223--1237.

\bibitem[{Speer et~al.(2017)Speer, Chin, and Havasi}]{Speer2017Conceptnet55}
R.~Speer, J.~Chin, and C.~Havasi. 2017.
\newblock {ConceptNet 5.5}: {A}n open multilingual graph of general knowledge.
\newblock In \emph{AAAI}.

\bibitem[{Stasaski and Hearst(2017)}]{Stasaski2017MultipleCQ}
K.~Stasaski and M.~A. Hearst. 2017.
\newblock Multiple choice question generation utilizing an ontology.
\newblock In \emph{BEA@EMNLP, 12th Workshop on Innovative Use of NLP for
  Building Educational Applications}.

\bibitem[{Sugawara et~al.(2017)Sugawara, Yokono, and
  Aizawa}]{Sugawara2016-rc-skills}
S.~Sugawara, H.~Yokono, and A.~Aizawa. 2017.
\newblock Prerequisite skills for reading comprehension: Multi-perspective
  analysis of mctest datasets and systems.
\newblock In \emph{AAAI}, pages 3089--3096.

\bibitem[{Trischler et~al.(2017)Trischler, Wang, Yuan, Harris, Sordoni,
  Bachman, and Suleman}]{Trischler2017-rc-newsqa}
A.~Trischler, T.~Wang, X.~Yuan, J.~Harris, A.~Sordoni, P.~Bachman, and
  K.~Suleman. 2017.
\newblock {NewsQA}: {A} machine comprehension dataset.
\newblock In \emph{Proceedings of the 2nd Workshop on Representation Learning
  for NLP}, pages 191--200.

\bibitem[{Turney(2017)}]{Turney2017LeveragingTB}
P.~D. Turney. 2017.
\newblock Leveraging term banks for answering complex questions: A case for
  sparse vectors.
\newblock \emph{CoRR}, abs/1704.03543.

\bibitem[{Weissenborn et~al.(2017)Weissenborn, Wiese, and
  Seiffe}]{Weissenborn2016-FastQa}
D.~Weissenborn, G.~Wiese, and L.~Seiffe. 2017.
\newblock Making neural qa as simple as possible but not simpler.
\newblock In \emph{CoNLL}, pages 271--280.

\bibitem[{Welbl et~al.(2018)Welbl, Stenetorp, and
  Riedel}]{QAngaroo:Welbl2017ConstructingDF}
J.~Welbl, P.~Stenetorp, and S.~Riedel. 2018.
\newblock Constructing datasets for multi-hop reading comprehension across
  documents.
\newblock \emph{TACL}.

\bibitem[{Zhang et~al.(2018)Zhang, Dai, Toraman, and Song}]{arc-kg2-Zhang2018}
Y.~Zhang, H.~Dai, K.~Toraman, and L.~Song. 2018.
\newblock {KG{\^{}}2: Learning to Reason Science Exam Questions with Contextual
  Knowledge Graph Embeddings}.
\newblock In \emph{arXiv}.

\end{thebibliography}

\clearpage
\appendix
\section{Knowledge Retrieval Module}
\label{appendix:retrieval}

\revtwo{This module is the first part of a two stage model for incorporating knowledge from an external source $K$. For each instance $(q,C)$ in the dataset, where $q$ is a question and $C = \{c_1, \ldots, c_4\}$ a set of answer choices, it performs information retrieval (IR) on $K$ to select a fixed size subset $K_{q, C}$ of potentially relevant facts.}
The second module is a neural network that takes $(q, C, K_{q, C})$ as input, and predicts the answer $a_{q, C}$.

For the IR module, we use TfIdfVectorizer\footnote{Term frequency, Inverse document frequency based vectorizer from {\em scikit-learn} \cite{scikit-learn}.} to build vector representations $\vec{q_{\tfidf}}$, $\vec{c^i_{\tfidf}}$ and $\vec{k_{\tfidf}}$ for the question $q$, choice $c_i \in C$, and fact $k \in K$ based on the tokens in the training set. We then calculate similarity scores $t_{q, k}$ and $t_{q, c_i, k}$ between $q$ and $c_i$, resp., and each of the external facts in $k \in K$:
\begin{align*}
\nonumber t_{q, k} &= 1- \similarity(\vec{q}_{\tfidf}, \vec{k}_{\tfidf})\\
t_{q, c_i, k} &= 1 - \similarity(\vec{c}^i_{\tfidf}, \vec{k}_{\tfidf}) \cdot t_{q, k},
\end{align*}
where $\similarity$ is implemented as cosine distance.
Based on these similarity scores, we obtain a set $K_{q, C}$ of facts for each $(q,C,K)$ as $K_{q} \cup \bigcup_i K_{q, c_i}$,
where $K_{q}$ and $K_{q,c_i}$ are the top $N_{k}$
facts each with highest similarity $t_{q, k}$ and $t_{q, c_i, k}$, \revtwo{respectively. $N_k$ is a hyper-parameter chosen from $\{5, 10, 20\}$ so as to yield the best \devname set performance.}

For experimentation with knowledge, we consider the `open book' set of facts $\mathcal{F}$ in conjunction with 
two sources of 
common knowledge: 
\revtwo{
the Open Mind Common Sense \cite{Singh2002-common-sense-kw-omcs} part of ConceptNet~\cite{Speer2017Conceptnet55}, and its WordNet~\cite{miller1995wordnet} subset.}

\section{Implementation and Training}
\label{appendix:implementation}

Our neural models are implemented with \textit{AllenNLP}\footnote{\url{https://allennlp.org}} \cite{Gardner2017AllenNLP} and \textit{PyTorch}\footnote{\url{https://pytorch.org}} \citep{paszke2017automatic}. 
We use \textit{cross-entropy} loss and the \textit{Adam} optimizer \cite{Kingma2015-adam} with initial learning rate 0.001. \revtwo{For the neural models \emph{without} external knowledge, we typically train the model with a maximum of 30 epochs and stop training early if the \devname set accuracy does not improve for 10 consecutive epochs. We also halve the learning rate if there is no \devname set improvement for 5 epochs. For the neural models \emph{with} external knowledge, we typically train for 60 epochs with a patience of 20 epochs.}
For most of our neural models, we use $h=128$ as the \textit{LSTM} hidden layer size. The embedding dropout rate is chosen from $\{0.1, 0.2, 0.5\}$, again based on the best \devname set performance. 

For each model configuration, we perform 5 experiments with different random seeds. For each run, we take the model with the best performance on \devname and evaluate on \testname. We report the average accuracy for the best \devname score and the average of the corresponding \testname score $\pm$ the standard deviation across the 5 random seeds.

\revtwo{The code for the models and the configuration files required for reproducing the results are available at {\small \dataurl}.}

\section{Additional Experiments}
\label{appendix:additional-experiments}
\subsection{Question Answering: ARC}
We also perform experiments with the  \textbf{Question Match} system on the Challenge (hard) set of the AI2 Reasoning Challenge or ARC~\cite{ARCClark2018}. We train several models with different LSTM hidden sizes (128, 256, \textbf{384 (best)}, 512), and dropout of the embedding layer (\textbf{0.0 (best)}, 0.2, 0.5) on the questions from the Challenge Train set and take the model that has the highest accuracy on the Dev set. The resulting system scores 33.87\% on the Challenge Test set, which is 2.17\% higher than the previous best score by \citet{arc-kg2-Zhang2018}. 
The code and model configuration are available at {\small \url{https://github.com/allenai/ARC-Solvers}}.
\subsection{Textual Entailment: SciTail}
We perform textual entailment experiments on the Science enTailment dataset SciTail~\cite{Khot2018Scitail}.
We change the \textbf{Question Match} model to a classic \textbf{BiLSTM Max-Out} \cite{Conneau2017} for textual entailment, by replacing the question $q$ and a choice $c_i$ with the premise $p$
 and the hypothesis $h$, resp., and perform binary classification on the entailment labels (Entail, Neural). We run experiments with BiLSTM encoders with LSTM hidden size of 384 and share the encoder parameters between the premise and the hypothesis. Without additional hyper-parameter tuning, this yields entailment accuracy scores of 87.9\% and 85.4\% on the Dev and Test sets, respectively.
 
\section{Success and Failure Examples}
\label{appendix:example-questions}
We give some examples of questions that were answered correctly/incorrectly by various groups of models. We include here the first three questions in each case.

\subsection{Neural Baseline Successes}

We begin with three examples of questions that all neural models
without external knowledge (namely Question Match, Plausible Answer, One-Odd-Out, and ESIM from the fourth group in Table \ref{table:question-ex-no-knowledge-correct})
predicted correctly.

\begin{table}[ht]
\centering
\vspace{-0.5cm}
\begin{tabular}{@{ }p{0.47\textwidth}@{}}
\\\hline
\question{A body may find its temperature to be lowered after} (A) water is heated up (B) \correct{fluid spreads from pores} (C) the air becomes arid (D) the sky stays bright 
\\\hline
\question{Oil is a non-renewable resource which tells us that when} (A) it can be remade (B) it can be found in other places (C) there is an endless supply (D) \correct{the final barrel is gone, there supply is finished} 
\\\hline
\question{Magma contains} (A) \correct{particles of iron} (B) Loads of leaves (C) Soda (D) Silly Putty 
\\\hline
\end{tabular}
\caption{Sample \question{questions} predicted \correct{correctly} (172/500) by all trained neural models without external knowledge.}   
\label{table:question-ex-no-knowledge-correct}
\vspace{-0.3cm}
\end{table}

In these examples, we observe that the correct answer usually contains a word that is semantically closer (than words in other answer choices) to an important word from the question: \textit{pores} to \textit{body};  \textit{non-renewable} (negative sentiment) to \textit{gone, finished} (also negative sentiment); \textit{iron} to \textit{magma (liquid rock)}.




\subsection{Neural Baseline Failures, Oracle Success}

\begin{table}[ht]
\centering
\begin{tabular}{@{ }p{0.47\textwidth}@{}}
\\\hline
\question{Frilled sharks and angler fish live far beneath the surface of the ocean, which is why they are known as} (A) \correct{Deep sea animals} (B) fish (C) Long Sea Fish (D) Far Sea Animals. 
\textbf{Oracle facts:} 
($f$) deep sea animals live deep in the ocean.
($k$) Examples of deep sea animals are angler fish and frilled sharks.
\\\hline

\question{Gas can fill any container it is given, and liquid} (A) is standard weight and size (B) is the opposite of variable (C) only needs a few (D) \correct{uses what it needs}.
\textbf{Oracle facts:} 
($f$) Matter in the liquid phase has definite volume.
($k$) liquid cannot spread endlessly.
\\\hline

\question{When birds migrate south for the winter, they do it because} (A) \correct{they are genetically called to} (B) their children ask for them to (C) it is important to their happiness (D) they decide to each year.
\textbf{Oracle facts:} 
($f$) migration is an instinctive behavior.
($k$)  instinctive is genetic.
\\\hline

\end{tabular}

\caption{Sample \question{questions} predicted \correct{correctly} by the $f+k$ Oracle model (405/500) but were predicted \incorrect{incorrectly} by all of the 4 neural models without knowledge (total of 69 out of 405). }   
\label{table:question-ex-oracle-f-k-correct-others-incorrect}
\vspace{-0.3cm}
\end{table}

\begin{table}[ht]
\centering
\begin{tabular}{@{ }p{0.47\textwidth}@{}}
\\\hline

\question{An example of data collection is:} (A - 0.9977) \incorrect{Deleting case files on the computer}, (B - 0.0000) Touching evidence without gloves, (C - 0.0004) \correct{speaking with a witness}, (D - 0.0019) Throwing documents in the trash. 
\textbf{Oracle facts}:
($f$) An example of collecting data is measuring.
($k$) Interviews are used to collect data.
\\\hline

\question{If a farmland up the hill gets rainfall, what could happen to lower lands?} (A - 0.0005) \correct{all of these}, (B - 0.0245) they could get fertilizer washed to them, (C - 0.9542) \incorrect{they could experience unfavorable chemical change in their lands}, (D - 0.0208) they could have their lands poisoned.
\textbf{Oracle facts}:
($f$) runoff contains fertilizer from cropland.
($k$) fertilizers for certain crops could poison other crops or soil types.
\\\hline

\question{Layers of the earth include all but:} (A - 0.0429) mantle, (B - 0.0059) \correct{center}, (C - 0.0334) crust, (D - 0.9177) \incorrect{inner core}.
\textbf{Oracle facts}:
($f$) the crust is a layer of the Earth.
($k$) the last layer is the outer core.
\\\hline

\end{tabular}

\caption{Sample \question{questions} predicted \incorrect{incorrectly} by all models models w/o knowledge, as well as the $f+k$ Oracle model, even though the Oracle model has confidence higher than 0.90.}   
\label{table:question-ex-oracle-f-k-incorrect-but-high-confidence}
\end{table}

Table \ref{table:question-ex-oracle-f-k-correct-others-incorrect} shows example questions (with the Oracle facts) from the Dev set that were predicted correctly by the $f+k$ Oracle model (405/500) but incorrectly by all of the 4 neural models without knowledge (69/405). In contrast to Table \ref{table:question-ex-no-knowledge-correct}, a simple semantic similarity is insufficient. The questions require chaining of multiple facts in order to arrive at the correct answer.

\subsection{Neural Baseline and Oracle Failures}

42/500 questions in the Dev set were predicted incorrectly by all models without external knowledge, as well as by the Oracle $f+k$ model. In Table \ref{table:question-ex-oracle-f-k-incorrect-but-high-confidence} we show 3 such questions. In all cases, the Oracle $f+k$ model made an incorrect prediction with confidence higher than 0.9. 

As noted earlier, there are several broad reasons why even this so-called oracle model fails on certain questions in OpenBookQA. In some cases, the core fact $f$ associated with a question $q$ isn't actually helpful in answering $q$. In many other cases, the corresponding second fact $k$ is noisy, incomplete, or only distantly related to $q$. Finally, even if $f$ and $k$ are sufficient to answer $q$, it is quite possible for this simple model to be unable to perform the reasoning that's necessary to combine these two pieces of textual information in order to arrive at the correct answer.

In the shown examples, the first question falls outside the domain of \textit{Science} where most of the core facts come from. The scientific fact ``($f$) An example of collecting data is measuring'' is transformed into a question related to the law and judicial domain of \textit{collecting data for a (court) case}. This is an indication that the model trained on the Train set does not perform well on distant domains, even if the core facts are provided.

In the second question, we have an option \textit{all of these}. Indeed, the selected answer seems the most relevant (a generalized version of the other two), but the model did not know that if we have an option \textit{all of these} and all answers are plausible, it should decide if all answers are correct and not pick the ``most likely'' individual answer.

The third question again requires the model to select a special type of aggregate answer (``all but xyz''), but the related Oracle facts are pointing to a specific answer.

\end{document}